\documentclass[10pt,twocolumn,letterpaper]{article}

\usepackage{wacv}
\usepackage{times}
\usepackage{epsfig}
\usepackage{graphicx}
\usepackage{amsmath}
\usepackage{amssymb}
\usepackage{import}
\usepackage{authblk}

\def\wacvPaperID{975} 

\wacvfinalcopy 

\ifwacvfinal
\def\assignedStartPage{1} 
\fi

\ifwacvfinal
\usepackage[breaklinks=true,bookmarks=false]{hyperref}
\else
\usepackage[pagebackref=true,breaklinks=true,colorlinks,bookmarks=false]{hyperref}
\fi

\ifwacvfinal
\else
\fi

\begin{document}

\title{FlowCaps: Optical Flow Estimation with Capsule Networks For Action Recognition}

\author[]{Vinoj Jayasundara}
\author[]{Debaditya Roy}
\author[]{Basura Fernando}
\affil[]{SCC, IHPC, A*STAR, Singapore.}

\maketitle

\begin{abstract}
Capsule networks (CapsNets) have recently shown promise to excel in most computer vision tasks, especially pertaining to scene understanding. In this paper, we explore CapsNet's capabilities in optical flow estimation, a task at which convolutional neural networks (CNNs) have already outperformed other approaches. We propose a CapsNet-based architecture, termed FlowCaps, which attempts to a) achieve better correspondence matching via finer-grained, motion-specific, and more-interpretable encoding crucial for optical flow estimation, b) perform better-generalizable optical flow estimation, c) utilize lesser ground truth data, and d) significantly reduce the computational complexity in achieving good performance, in comparison to its CNN-counterparts. 

\end{abstract}

\section{Introduction}
Optical flow represents apparent motion of objects, surfaces, and edges in a visual scene caused by the relative motion between an observer and a scene~\cite{Horn1981}.
Given a pair of images from the scene, optical flow estimates the displacement of pixels in spatial domain.
Optical flow is important for many applications, including action recognition, motion detection, tracking and autonomous driving.
In the recent years, convolutional neural networks (CNNs) have made breakthroughs in a variety of computer vision tasks, including optical flow estimation~\cite{dosovitskiy2015flownet,Ranjan2017,Hui2018,sun2018pwc,yin2018geonet,zhu2017deep,bar2020scopeflow}. 
For example, FlowNet is an end-to-end trainable CNN to solve the optical flow estimation problem in a data-driven, supervised fashion, which outperforms the conventional curated-feature driven models such as~\cite{Revaud2015}.
Ideally, deep optical flow estimation methods should be equivariant which allows us to obtain feature representation equivalent to geometric changes in the image space.
This would allows us to obtain accurate optical flow estimation by measuring feature displacements in functional form using deep neural networks.
Despite the success of CNNs and CNN-based optical flow estimation, they suffer from the issue of invariance to certain geometric attributes such as translation and affine changes.
On the other-hand capsule networks (CapsNets) \cite{hinton2011transforming,sabour2017dynamic} marked a milestone by identifying and attempting to resolve several key limitations of CNNs, such as the inability to understand spatial relationships between features, being invariant rather than equivariant, inept routing of data between the layers, among others.   
Therefore in this paper we exploit capsule networks for optical flow estimation task, and to explore the aptness and potential gains caused by the resulting optical flows in auxiliary tasks such as action recognition.

Precise optical flow estimation requires pixel-wise localization as well as correspondence matching between frames. However, raw pixel intensities from a pair-of-frames carry sparse motion-related information useful for optical flow estimation, often cluttered with non-motion related information. 
Hence, a key sub-task of an optical flow estimator is to successfully untangle motion-related information from raw pixel intensities. Yet, this sub-task is specially challenging for a CNN, since they learn primarily invariant encoding comprising of high level entities, and often lose other useful internal information pertaining to the pose, orientation, whole-part relationships, and other physical properties of such entities, along with the spatial relationships between them. 
Furthermore, discarding information that are not useful to the task at hand at a lower level, prior to passing them to higher levels is crucial to the untangling process. 
Yet, CNNs fail to  filter out such unnecessary information since they routes data between layers using pooling operators, especially during the initial training iterations. 
Hence, we argue that the optical flow estimation task will heavily benefit from a more comprehensive and selective encoding mechanism provided by capsule networks. 

Capsule networks excel at comprehensively encoding the physical properties of the entities present in the inputs within their instantiation parameter vectors, while learning the part-whole spatial relationships between such entities. 
Studies have shown that the physical properties captured by the capsules are often relevant to the task at hand, especially the properties corresponding to the instantiation parameters with the highest variances, and that the encoding learnt by capsules are highly interpretable, by means of a post-training perturbation analysis \cite{jayasundara2019textcaps}. 
The same observation can be extended in our case to assume that the representation learnt by a capsule encoder will comprise motion-specific properties useful for optical flow estimation. Hence, similar entities in the input images receive similar encoding with finer-grained representations than CNNs, allowing the correspondence-matching to be more convenient and precise. 
Furthermore, the dynamic routing algorithm \cite{sabour2017dynamic} deployed in capsule networks achieves coincidence filtering, which arguably aids the untangling process. 
Higher level capsules represent complex motion-related entities with higher degree of freedom, and dynamic routing ensures that lower level entities that have little agreement with these (non-motion related entities) are effectively cut-off from the forward propagation. 
Hence, we hypothesize that the use of a capsule encoder will aid the optical flow estimation task by providing finer-grained, motion-specific and more-interpretable encoding, in comparison to its CNN counterpart.

Optical flow estimation is an object or class agnostic task. The specific identities and the categories of the objects and the actors are not attributed to the task, only where and what kind of motion take place instead. 
Hence, it is intuitive that, if the meta-level conditions (such as the presence of camera motion, range of displacement, and etc.) do not drastically change, optical flow estimation models should generalize beyond the data that they are trained for. 
To further explore this property, among other reasons, we consider action recognition as an auxiliary task. 
More specifically, we compare the capabilities of CNN and capsule networks in estimating class-agnostic optical flows, and generalizing to other classes when trained on a subset of action classes. 
CNNs are generally translation invariant, and not invariant to other transforms such as the orientation changes. 
Hence, they require a lot of training data with ample variations to learn to handle such transforms, resulting in reduced generalization capabilities. 
In contrast, capsule networks are equivariant, where lower level capsules exhibit place-coded equivariance and higher level capsules exhibit rate-coded equivariance \cite{sabour2017dynamic}. For instance, CNNs learn rotational invariance by training on a large number augmented images, whereas capsule networks learn to encode rotation in their instantiation parameters without observing many such augmentations. 
Subsequently, capsule networks are able to successfully encode rotation, even for an image outside the training domain. Hence, we argue that capsule networks will better-generalize to unseen action classes for the optical flow estimation, and require comparatively lesser ground truth data with fewer augmentations to achieve similar performances as CNNs.

In addition, capsule encoders provide a low-dimensional concise representation in comparison to shallow convolutional feature maps, and they undertake a significant portion of the burden of untangling motion-related information. 
Hence, it reduces the workload from network that generates optical flow image known as expanding network. 
Simultaneously, a capsule encoder itself has less number of trainable parameters than its direct CNN-counterpart, as capsule networks group neurons together yielding in a fewer number of connections between layers. Hence, a capsule encoder contributes to drastically reducing the computational complexity of the overall.

Estimated optical flows have a wide-utility in a range of computer vision tasks, and action recognition is one such task \cite{simonyan2014two}. 
It is well known that motion stream obtained via optical flow is complimentary to the spatial stream.
As a downstream task, we experiment with action recognition using the estimated optical flows from our models. 
We investigate two key approaches for this task, the standard frame-wise approach \cite{gao2018im2flow} as well as a segment-wise approach which considers a set of consecutive frames together for optical flow estimation (in contrast to two consecutive frames in the frame-wise approach), in an attempt to benefit from the additional contextual information in the segments.
We demonstrate that segment-wise optical flow estimation with our model is more accurate and obtains better action recognition results.
To this end, we propose a capsule networks-based architecture for optical flow prediction and activity recognition, leveraging on the dynamic routing algorithm \cite{sabour2017dynamic}. More specifically, we make the following contributions in this paper.

First, we propose a capsule networks based architecture, termed FlowCaps, to achieve better optical flow estimation than its convolutional counterpart. To the best of our knowledge, this is the first attempt to investigate the use of capsules for this task. Furthermore, we utilize the estimated optical flows for action recognition, and propose a modified loss function that improve upon the existing EPE loss.
    
Second, we evaluate the performance of FlowCaps model on several datasets where we outperform other baselines in both optical flow estimation and action recognition while being less computationally complex. Furthermore, we investigate the capabilities of FlowCaps in terms of out-of-domain generalization and training with only a few samples, in comparison to baselines. 

\section{Background and Related Works}
The concept of grouping neurons to form a capsule was first proposed by Hinton \textit{et al.} in \cite{hinton2011transforming} and extended by Sabour \textit{et al.} in \cite{sabour2017dynamic} introducing the dynamic routing algorithm to route sets of capsules between layers. 
These models are primarily used for image classification, image parameterization and image reconstruction.
Capsule networks have proven that they excel at various computer vision tasks throughout the literature. CapsGan\cite{jaiswal2018capsulegan} utilized capsule networks as a discriminator which produced visually better results than conventional CNN-based GANs. 
Moreover, SegCaps\cite{lalonde2018capsules} implemented a capsule network based architecture for image segmentation and was able to achieve state-of-the-art results on datasets such as LUNA16. 
Further, Zhao \textit{et al.}\cite{zhao20193d} employed capsule networks to classify, reconstruct and perform part-based segmentation on sparse 3D point clouds. 
Extending capsule networks into video analysis, Duarte \textit{et al.} \cite{duarte2018videocapsulenet} introduced VideoCapsuleNet which consists of convolutional capsule layers and capsule pooling layers in order to facilitate action recognition.
Our method is drastically different from these models as we use capsule network encoder to obtain a representation suitable for optical flow estimation and then use a so called expanding network to generate optical flow images.
We also modified the capsule network architecture to avoid issues related to squashing function and make architectural changes to cater for optical flow estimation.


Deep nets based optical flow estimation has been studied in FlowNet, FlowNet 2.0, SpyNet and LiteFlowNet. \cite{dosovitskiy2015flownet,ilg2017flownet,Ranjan2017,Hui2018}.
LiteFlowNet is composed of two compact sub-networks that are specialized in pyramidal feature extraction and optical flow estimation~\cite{Hui2018}. 
This method is able to extract features faster compared to~\cite{dosovitskiy2015flownet}.
Similarly, SPyNet~\cite{Ranjan2017} also uses spatial pyramids similar to~\cite{Hui2018} with a compact network.
Spatial pyramids are used to make sure that the representation captures some global spatial information.
However, our method is able to preserve physical properties in the image space including the spatial structure of entities due to properties of capsule networks.
Therefore, we do not need to use a multi-scale approach.
Some methods also use additional tools such as external edge detectors or image patch-based correlations in estimating optical flow.
Author in~\cite{Zweig2017} interpolates third-party sparse flows using a off-the-shelf edge detector. 
DeepFlow~\cite{Weinzaepfel2013} uses convolution and pooling operations similar to traditional CNNs, however the filter weights are non-trainable image patches. 
In-fact, similar to FlowNet, DeepFlow also uses correlation.
EpicFlow~\cite{Revaud2015} uses externally matched flows as initialization and then performs interpolation. 
Similarly, Im2Flow \cite{gao2018im2flow} and Selflow \cite{liu2019selflow} are related to us.
However, to the best of our knowledge, we are the first to use capsule network-based architecture for optical flows estimation which in principle is a better choice than traditional CNNs for this task.

Action recognition methods have benefited a lot from optical flow~\cite{simonyan2014two} and other methods use dense optical-flow obtained by motion hallucination for action recognition~\cite{gao2018im2flow}.
Some methods such as Dynamic Images \cite{Bilen2016,Bilen2017} generate motion images for action recognition using rank pooling~\cite{Fernando2015,Fernando2016}.
Motion images are even used for tasks such as still image action recognition~\cite{herath2019} and action anticipation~\cite{rodriguez2018}.
In this work we also use action recognition as the primary application of optical flow estimation, nevertheless, action recognition is not the primary focus of this paper.

\section{FlowCaps: Network Architecture}
\begin{figure}[!h]
\centering
\includegraphics[width=\columnwidth]{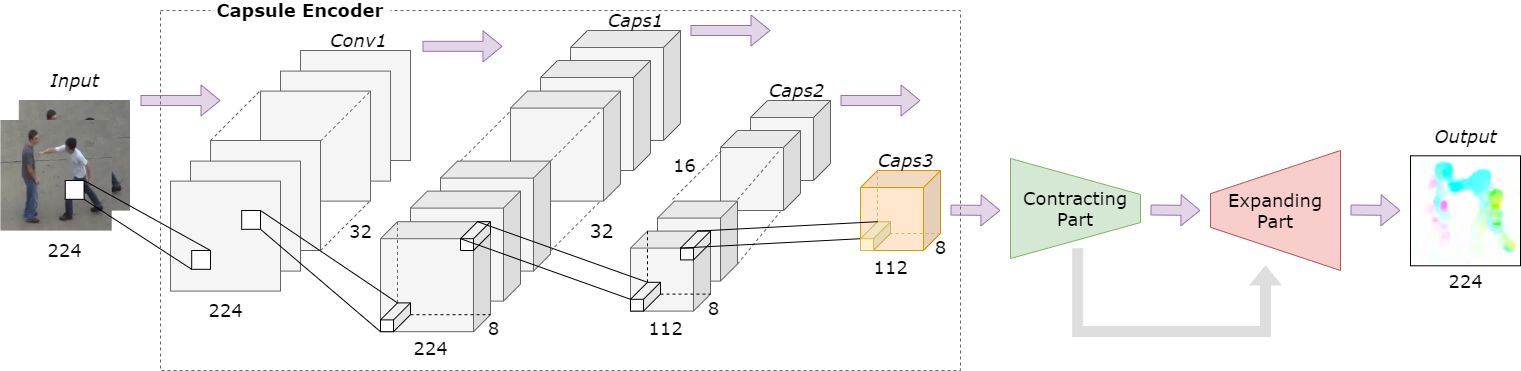} 
\caption{The FlowCaps architecture. The input is fed to the capsule encoder, which passes a concise representation of the input to the subsequent contracting and expanding parts, for optical flow estimation.}
\label{fig:flowcaps}
\end{figure}

Authors of \cite{dosovitskiy2015flownet} proposed the first end-to-end trainable CNN architecture termed FlowNetS for optical flow estimation. 
Motivated by the benefits of both FlowNetS and capsule networks, in this work we propose a new model for optical flow estimation by borrowing concepts from both FlowNetS and capsule networks.
We call our model FlowCaps-S.
%
%
Given a pair of RGB images, FlowCaps-S estimate optical flows using the architecture described in section~\ref{sec:flowcaps_arch}.
If the input is a video, then we show how to use generated optical flow to perform action recognition in section~\ref{sec:action_arch}.

\subsection{FlowCaps-S Architecture} 
\label{sec:flowcaps_arch}

In FlowNet, the network learns the optical flow estimation task by applying convolutional filters on the raw pixel values of the considered image pairs. 
However, extraction of motion information becomes more difficult, especially when the datasets have small realistic displacements. 
We hypothesize that the use of a capsule encoder as illustrated in Fig. \ref{fig:flowcaps} instead of a shallow convolutional encoder prior to the contracting part similar to FlowNet, learns a finer-grained, concise, and more interpretable representation of the physical properties attributed to motion, by eliminating the information unrelated for motion via dynamic routing \cite{sabour2017dynamic}.
Furthermore, our FlowCaps-S benefit from equivariant properties as outlined in the introduction.
Now we describe the details of our FlowCaps-S model.  

As illustrated in Fig. \ref{fig:flowcaps}, the proposed capsule encoder consists of a convolutional layer (\textit{Conv1}) having 32 kernels of size 7 $\times$ 7 and a Leaky ReLU activation, followed by three convolutional capsule layers coined \textit{Caps1}, \textit{Caps2} and \textit{Caps3}. 
The \textit{Caps1} layer consists of 32 channels of 8-dimensional capsules, which will be dynamically routed to each of the 16 channels of 8-dimensional capsules in the \textit{Caps2} layer. 
We adopt the dynamic routing algorithm proposed in \cite{sabour2017dynamic}, with the exception of squashing of capsule output vectors. 
The squash function is used to ensure that the length of each capsule output is kept between $0-1$, as the length represents the probability of existence. 
Albeit existence of motion is crucial for optical flow prediction, we do not utilize the probabilities in a mathematical sense (for instance as in classification), and hence do not require the output lengths to be kept between $0-1$. 
On the other hand, squashing high dimensional vectors leads to issues such as extremely small individual values and vanishing gradients. 
Hence, we do not utilize the squash function. 
Subsequently, the representation is further projected down spatially as well as feature-wise, via dynamic routing to a single channel of 8-dimensional capsules in the \textit{Caps3} layer. 
The decision to keep the output of the capsule encoder to one channel stems from the capsule representational assumption. 
Hence, at each location in the input image, there is at most one instance of the type of entity that a capsule represents. 
This allows us to have a single capsule channel in our representation.
This is useful for optical flow estimation task.


Let $\Psi^{l} \in \mathbb{R}^{ (h_1 \times w_1 \times c_1 \times n_1)} $ and $\Phi^{l}  \in \mathbb{R}^{ (h_2 \times w_2 \times c_2 \times n_2)} $ be the input and output of the capsule layer $l$, where $h,w$ are the spatial dimensions, and $c,n$ are the number of channels and the dimensionality of each capsule respectively. 
Here, $\Phi^{l-1} \equiv \Psi^{l}$, and  $\psi^l_i$ is the $i^{th}  (i \in [1,h_1 \times w_1 \times c_1]$) capsule in layer $(l-1)$ and $\phi^{l}_{j}$ is the  $j^{th} (j \in [1,h_2 \times w_2 \times c_2]$) capsule in the layer $l$. 
First, $\Psi^{l}$ is reshaped in to $(h_1,w_1, c_1\times n_1)$ to prepare the channels for the convolution operation, and subsequently convolved with $(c_2 \times n_2)$ filters producing an output of the shape $(h_2,w_2,c_2\times n_2)$, which is then reshaped to $(h_2,w_2,c_2,n_2)$. Subsequently, each $\psi^l_i$ is routed to each $\phi^{l}_{j}$ dynamically based on their agreement $a_{ij} = \widehat{\psi}^l_i \cdot \phi^{l}_{j}$, where $\widehat{\psi}^l_i = W_{ij}\psi^l_i$ and $W_{ij}$ is the trainable transformation matrix which projects $\psi^l_i$ from its native space to the higher dimensional space of $\phi^{l}_{j}$.

\begin{figure}[!h]
\centering
\includegraphics[width=\columnwidth]{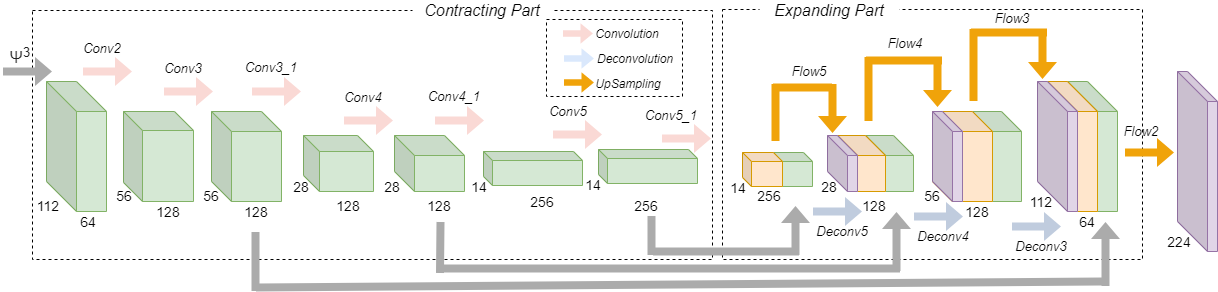} 
\caption{ The FlowCaps contracting and the expanding parts. The output of the capsule encoder, $\Psi^{3}$, is fed in to the contracting part, followed by the expanding part, which estimates the optical flows.}
\label{fig:conex}
\end{figure}

The reduced representation  $\Psi^{3}$ is then fed to the contracting part followed by the expanding path, which are simplified versions of those proposed in FlowNet. The simplified contracting part comprises seven convolution blocks with batch normalization and Leaky ReLU activation. Downsampling by a factor of 2 occurs every other block, starting from the second, with hyperparameters as illustrated by Fig. \ref{fig:conex}. The resultant feature map is passed on to the expanding path, which comprises four blocks of upsampling. Each block constitutes a deconvolution layer with Leaky ReLU activation which upsamples the feature maps by a factor of 2, followed by a concatenation with the corresponding block skip connected from the contracting part. Further, we use skip connections between the contracting and expanding parts to nourish the information flow and provide lower-level entity information to the deconvolutional layers, similar to FlowNet.

\subsection{Improvements to the Loss Function}

The loss function that is used in the state-of-the-art optical flow estimation approaches, endpoint error (EPE), sums the $L_2$ norms of the difference between the individual components of the ground truth and estimated flow fields. We identify two key issues of using EPE loss in optical flow estimation. First, EPE only considers the magnitude component of the vector field in its calculations, whereas the angle component is omitted. Yet, the angle component carries important information helpful for the optical flow estimation task. Second, the $L_2$ norm is highly susceptible to outliers with higher values, even a few can have a significant impact on the loss value. In an attempt to alleviate these key issues, we propose the following loss function,
\begin{equation} \label{eq:newloss}
    L = L_{mag} + \alpha L_{ang}
\end{equation}

\begin{align}
\label{eq:logcosh}
L_{mag} = & \frac{1}{N} \sum_{i=1}^N log(ln(\frac{e^{(u_i^p-u_i^t)}+e^{(u_i^t-u_i^p)}}{2})\\ \nonumber
+ & ln(\frac{e^{(v_i^p-v_i^t)}+e^{(v_i^t-v_i^p)}}{2}))
\end{align}
\begin{equation} \label{eq:cosine}
    L_{ang} = \frac{1}{N} \sum_{i=1}^N (1-\frac{u_i^pu_i^t+v_i^pv_i^t}{\sqrt{(u_i^p)^2+(v_i^p)^2}\sqrt{(u_i^t)^2+(v_i^t)^2}})||\mathbf{T_i}||_2
\end{equation}
where $\alpha$ is an empirically determined constant, $N$ is the mini-batch size, $u_i^p,u_i^t,v_i^p,v_i^t$ are the respective $u,v$ components of the estimated ($p$) and ground truth ($t$) optical flows, and $||\mathbf{T_i}||_2 = \sqrt{(u_i^t)^2+(v_i^t)^2}$. We propose the logcosh loss function for the magnitude loss $L_{mag}$, denoted in equation \ref{eq:logcosh}, since it is more robust to outliers while behaving similar to the $L_2$ loss. Furthermore, we propose a variation of the cosine similarity for the angular loss, which adaptively scales the loss with respect to the magnitude  of the ground truth optical flows. The scaling aids to alleviate the issue of undefined loss values and gradients when both ground truth and the predicted fields are zero vectors.

\subsection{Optical flow estimation for activity recognition}
\label{sec:action_arch}
In this study, we attempt two fundamental computer vision tasks, namely, optical flow estimation followed by activity recognition. To this end, we consider two different approaches based on the number of consecutive frames ($k$) considered for prediction at a time. First, frame-wise prediction focuses on estimating optical flows from only a pair of consecutive frames ($k=2$), similar to \cite{dosovitskiy2015flownet}. Subsequently, the estimated optical flows are utilized in action recognition, producing an action label per the said pair of images. Second, segment-wise prediction focuses on simultaneously estimating the optical flows for a whole segment of consecutive frames ($k>2$). Similarly, the action recognition is performed on the set of estimated optical flows, producing an action label per the said segment.

We hypothesize that in the datasets where motion information are predominant, action classification performed with optical flows achieves similar performance as with original rgb frames, yet, requires shallower models that are much faster. 
In the case of datasets where static information are also significant, motion information derived from the optical flows can be combined with the static information to achieve similar performance, similar to~\cite{simonyan2014two}.

\subsection{Frame-wise Prediction} \label{sec:frame_pred}
The input to the frame-wise prediction network $\mathbf{X_{frm}} \in \mathbb{R}^{(H\times W\times 2C)}$ is stacked along the channel dimension, where $H,W$ denote the spatial dimensions and $C$ denotes the number of channels per frame. $\mathbf{X_{frm}}$ is fed to the network and the estimated optical flows $\mathbf{\hat{Y}_{frm}}$ are compared against the ground truth $\mathbf{Y_{frm}}\in \mathbb{R}^{(H\times W\times 2)}$. Note that optical flow images have two channels, the flows in x and y directions.
 Subsequently, the action recognition is performed with a shallow CNN using predicted optical flow images $\mathbf{\hat{Y}_{frm}}$ as the input. Our simple optical flow classification network consists of five blocks of convolutional layers and a maxpooling layer each, followed by two fully connected layers. The convolutional layers each have $3 \times 3$ filters with ReLU activation, whereas the fully connected layers have $32$ and $\kappa$ units with ReLU and softmax activation respectively, where $\kappa$ is the number of action classes.
 We train this network from scratch.

\subsection{Segment-wise Prediction}
Typically, the optical flow is estimated using consecutive pair of frames, however, models can benefit from additional contextual information.
One solution is to predict the optical flow for a given video segment consisting of more than two frames. 
To be precise, our segment-wise prediction considers $k$ number of frames to be stacked together as the input $\mathbf{X_{seg}} \in \mathbb{R}^{(k \times H\times W\times C)}$. 
We modify FlowNetS and FlowCaps-S models to handle segment-wise optical flow prediction using 3D convolutions. Specifically, we adopt concepts in I3D \cite{girdhar2019video} and DeepCaps \cite{rajasegaran2019deepcaps}. The modified FlowNetS-3D and FlowCaps-S-3D models estimate the optical flows $\mathbf{\hat{Y}_{seg}}$ which are compared against the ground truth optical flow $\mathbf{Y_{seg}}\in \mathbb{R}^{(H\times W\times 2)}$ corresponding to the middle two frames of the segment. Subsequently, action recognition is performed with $\mathbf{\hat{Y}_{seg}}$ similar to Section \ref{sec:frame_pred}.

\section{Experiments and Results}
\begin{table*}[t]
\begin{center}
\begin{tabular}{|l|l|c|c|c|c|}
\hline
\multicolumn{2}{|l|}{Model} & Params (M) & \ Sintel clean \ & \ Sintel final \ & \ KITTI15 \\ 
\hline\hline
Conventional & EpicFlow \cite{Revaud2015} & - & 2.27 & 3.56 & 9.27\\ 
& FlowFields \cite{bailer2015flow} & - & \textbf{1.86} & 3.06 & 8.33 \\
\hline
Heavyweight CNN & FlowNetS \cite{dosovitskiy2015flownet} & 38.68 & 4.50 & 5.45 & - \\
& FlowNet2 \cite{ilg2017flownet} & 162.49 & 2.02 & 3.54 & 10.08 \\
\hline
Lightweight CNN & LiteFlowNet \cite{Hui2018} & 5.37 & 2.48 & 4.04 & 10.39 \\
& SPyNet \cite{Ranjan2017} & 1.20 & 4.12 & 5.57 & - \\

& Ours & 2.39 & 2.13 &  \textbf{2.51} & \textbf{7.83} \\
\hline
\end{tabular}
\end{center}
\caption{Comparison of frame-wise training EPE values across different approaches for  optical flow estimation datasets.}
\label{tbl:epe_opflow}
\end{table*}

\begin{table*}[!h]
\begin{center}
\begin{tabular}{|l|cc|cc|cc|cc|}
\hline
Model  & \multicolumn{2}{c|}{UCF I-Frames} & \multicolumn{2}{c|}{UTI P-Frames} & \multicolumn{2}{c|}{KTH I-Frames} & \multicolumn{2}{c|}{JHMDB} \\
\hline\hline
& test epe & action & test epe & action & test epe & action & test epe & action \\
\cline{2-9}
GT & - & 79.4\% & - & 81.37\% & - & 68.90\% & - & 51.49\%\\
FlowNetS & 1.53 & 55.58\% & 0.44 & 84.12\% & 1.19 & 61.30\% & 0.49 & 44.03\%\\
LiteFlowNet & - & - & - & 83.17\% & - & 59.79\%& - & 40.30\%\\
SPyNet & \textbf{1.37} & \textbf{65.78\%} & 0.42 & 87.66\% & 0.95 & 64.30\% & 0.44 & 42.54\% \\
\hline 
Ours & 1.49 & 64.49\% & 0.39 & 86.02\% & 1.10 & 65.00\% & \textbf{0.40} & \textbf{48.51\%}\\
Ours - Mod Loss* & 1.41 & - & 0.35 & - & 1.04 & - & 0.26 & -\\
Ours - Segment & 1.40 & 65.16\% & \textbf{0.37} & \textbf{88.34\%} & \textbf{0.93} & \textbf{72.50\%} & 0.71 & 41.90\%\\
\hline
\end{tabular}
\end{center}
\caption{Comparison of frame-wise testing EPE values and action recognition performances by different approaches on the benchmark action datasets. *Training utilizing the modified loss function proposed in eq. \ref{eq:newloss}, and testing using the EPE loss for comparison.}
\label{tbl:epe_action}
\end{table*}

\begin{table}[t]
\begin{center}
\resizebox{\columnwidth}{!}{%
\begin{tabular}{|l|c|c|c|c|c|c|}
\hline
Model  & \multicolumn{2}{c|}{KTH-I Frames} & \multicolumn{2}{c|}{Sub UCF-I Frames} & \multicolumn{2}{c|}{UTI-P Frames}  \\
\hline\hline
&\multicolumn{6}{|c|}{Optical flow estimation performance in EPE}\\ \hline
& Frame & Seg.  & Frame  & Seg. & Frame  & Seg.\\
\cline{2-7}
FlowNetS & 1.1934 & 1.1355 & 2.3149 & 2.3079 & 0.4426 & 0.4265\\
FlowCaps-S & 1.1033 & \textbf{0.9384} & 2.2037 & \textbf{2.1930} & 0.3806 & \textbf{0.3672} \\
\hline
&\multicolumn{6}{|c|}{Action classification performance}\\ \hline
FlowNetS & 61.30\% & 66.30\% & 85.50\% & 89.70\% & 84.12\% & 83.08\%\\
FlowCaps-S & 65.00\% & \textbf{72.50\%} & 91.20\%  & \textbf{92.30\%} & \textbf{86.02\%} & 85.93\% \\
\hline
GT & \multicolumn{2}{c|}{68.90\%} & \multicolumn{2}{c|}{92.60\%} & \multicolumn{2}{c|}{81.37\%} \\ \hline
\end{tabular}
}
\end{center}
\caption{The frame-wise and segment-wise testing EPE values and classification performance achieved by FlowNetS and FlowCaps-S models on the KTH I-frames, Sub UCF I-frames, and UTI P-frames datasets.}
\label{tbl:epe}
\vspace{-4mm}
\end{table}

In this section we evaluate the validity of our method on several optical flow estimation and video action recognition datasets.
Following prior deep optical flow estimation methods~\cite{ilg2017flownet,Hui2018}, we use Sintel~\cite{Butler2012} and KITTI15~\cite{geiger2013vision} benchmarks for evaluation.
We also use the UCF101 \cite{soomro2012ucf101}, UTI~\cite{ryoo2010ut}  KTH~\cite{schuldt2004recognizing} and JHMDB~\cite{JhuangICCV2013} datasets for action recognition related experiments. 
More specifically, for optical flow estimation, we extract frames from the videos and stack $k$ ($k=2$ for frame-wise, and $k>2$ for segment-wise) consecutive frames together as the input to the network. However, initial experiments revealed that consecutive frames directly extracted from the videos contained little motion information, yielding trivial optical flows. As a solution, we extracted I-frames and P-frames, corresponding to the keyframes of the video in order to create the following datasets.\\

\noindent
\textbf{KTH I-Frames:} All the 6 action classes in the KTH dataset were used for training. After I-frames extraction, 19 videos with only one I-frame were removed, yielding 5,811 samples which are randomly split at a 8:2 ratio for training and testing respectively.

\noindent
\textbf{Sub UCF I-Frames:} We use the following five classes from the UCF 101 dataset: \textit{Rowing, BenchPress, CleanAndJerk, HulaHoop}, and \textit{Lunges}, selected at random based on the availability of sufficient (more than 5 per video) I-frames, amount of movement and the presence of camera motion. Extraction of I-frames on these five classes yielded in 10,262 optical flow samples which are randomly split at a 8:2 ratio for training and validation respectively.
Subsequently, we use I-frames extracted from the rest of the UCF-101 classes as the testing set for out-of-domain generalization.

\noindent
\textbf{UTI P-Frames:} All the 6 action classes in the UTI dataset were used for training. However, the extraction of I-frames yielded only 1 frame per video. Hence, we decided to use P-frames instead, which yielded in 2,748 optical flow samples, which are split according to the predefined groups as in \cite{ryoo2010ut} for training and testing.

\noindent
\textbf{Implementation details:}
We used PyTorch for the development of FlowCaps. All the optical flow estimation models and the action recognition models were trained on GTX-2080Ti and GTX-1080Ti GPUs for 1000 epochs and 50 epochs respectively, using the Adam optimizer with the learning rate set to 0.001. 
For the models trained on the above datasets, FlowNetS \cite{ilg2017flownet} has 38.68 million trainable parameters, whereas the proposed FlowCaps-S has only 2.39 million trainable parameters. This is a drastic reduction in the computational complexity with a significant margin of $94\%$ by FlowCaps-S, while surpassing the performance of FlowNetS. The reduction in computational complexity can be directly attributed to the hypothesized concise representation achieved by the capsule encoder.

\subsection{Evaluating optical flow estimation}
In this section we compare our FlowCaps-S model with other state of the art methods in the literature~\cite{Revaud2015,bailer2015flow,dosovitskiy2015flownet,ilg2017flownet,Hui2018,Ranjan2017}. 
We train all models in Singtel clean dataset and then evaluate the performance on other dataset using the same model.
For comparisons, we may classify prior methods into three categories; "conventional", "heavyweight" and "lightweight" CNN. 
Our method also falls under lightweight CNN category. Results are reported in Table~\ref{tbl:epe_opflow}. From the results, we can conclude that our method performs the best in lightweight category and only FlowFields \cite{bailer2015flow} method outperforms our results on Sintel clean dataset.
Our method outperforms recent methods such as LiteFlowNet \cite{Hui2018} across all compared datasets and our method has good out of domain generalization performance.
Our method attains good results due to the suitable properties of capsule networks for optical flow estimation task such as equivariance.
We conclude on conventional datasets, our method is able to outperform most of the other methods by a considerable margin for optical flow estimation utilizing relatively small amount of parameters.

\subsection{Evaluating on action recognition}
In this section we compare our FlowCaps model with other methods using four action recognition benchmarks for optical flow estimation and action recognition.
Results are reported in Table~\ref{tbl:epe_action}.
Overall our method obtains better results than other recent methods on several datasets.
JHMDB is a dataset having lot of motion and it benefits from motion stream. While the ground truth optical flow obtains a classification accuracy of 51.49\% our FlowCaps model obtains 48.51 outperforming all other state-of-the-art methods such as SPyNet.
Interestingly, on this dataset we obtain the best EPE score of 0.40 while SpyNet obtains only 0.44.
On UCF-I frames the ground truth optical flow obtains 79.4\% while our method obtains the only of 60.05\% where state-of-the-art SPyNet obtains better results than us.
We conclude that out model performs well on both traditional optical flow estimation benchmarks as well as on action recognition datasets. 

Furthermore, we investigate on the effect of the modifications proposed to the loss function in Equation \ref{eq:newloss}, with $\alpha$ empirically set to $0.1-0.2$. For a fair comparison, we train using the modified loss, while testing with the conventional EPE loss. We obtain significant improvements in optical flow estimation for all four datasets, as reported in Table \ref{tbl:epe_action} (Ours - Mod Loss*), establishing that the proposed modifications to the loss function are effective.  

\subsection{Evaluating the impact of segment-wise and frame-wise model}
\newcommand{\KTHTH}[1]{{\frame{\includegraphics[height=60pt,width=82pt]{figs/frame_kth/#1}}}}
\begin{figure}[t]
\centering
\resizebox{0.99\linewidth}{!}{
\begin{tabular}{cccc}
Image & Ground Truth & FlowNetS & FlowCaps-S\\
\KTHTH{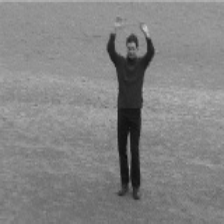} & \KTHTH{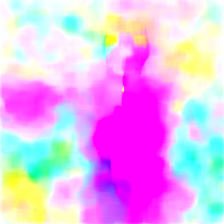} & \KTHTH{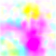} & \KTHTH{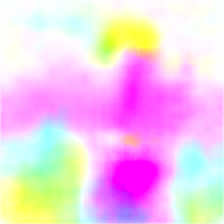}\\
\KTHTH{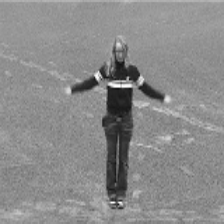} & \KTHTH{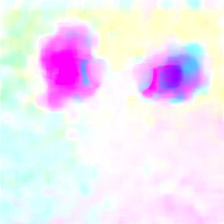} & \KTHTH{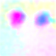} & \KTHTH{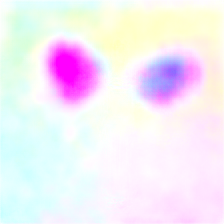}\\
\KTHTH{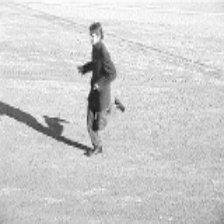} & \KTHTH{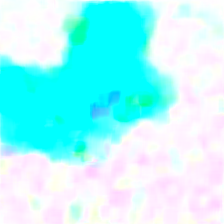} & \KTHTH{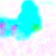} & \KTHTH{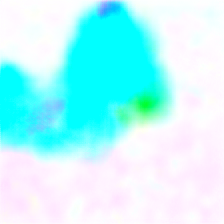}\\
\end{tabular}
}
\caption{The optical flow estimation results on the KTH I-Frames dataset.}
\label{fig:frame_kth}
\end{figure}
\newcommand{\UCFTH}[1]{{\frame{\includegraphics[height=60pt,width=82pt]{figs/frame_ucf/#1}}}}
\begin{figure}[!h]
\centering
\resizebox{.99\linewidth}{!}{
\begin{tabular}{cccc} 

Image & Ground Truth & FlowNetS & FlowCaps-S\\
\UCFTH{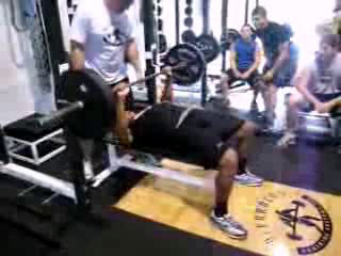} & \UCFTH{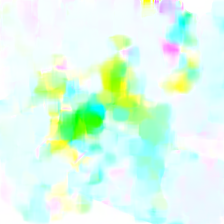} & \UCFTH{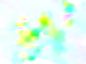} & \UCFTH{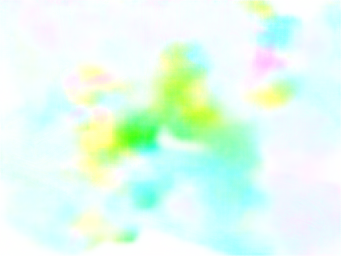}\\
\UCFTH{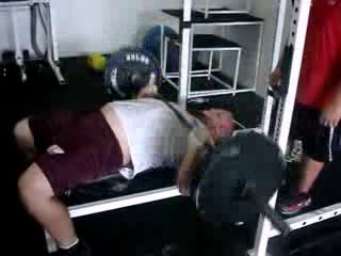} & \UCFTH{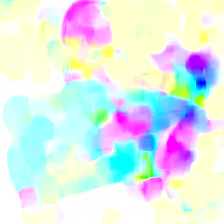} & \UCFTH{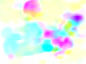} & \UCFTH{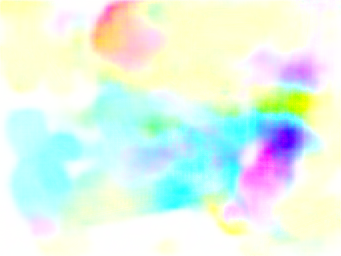}\\
\UCFTH{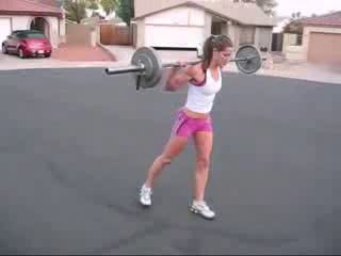} & \UCFTH{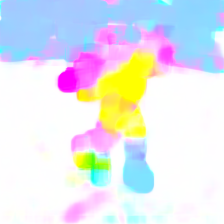} & \UCFTH{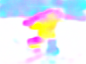} & \UCFTH{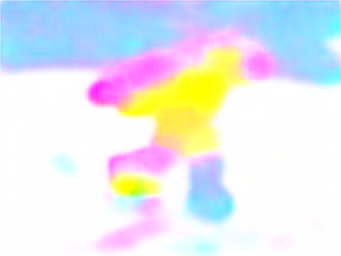}\\
\end{tabular}
}
\caption{The optical flow estimation results on the UCF I-Frames dataset.}
\label{fig:frame_ucf}
\end{figure}
\newcommand{\UTITH}[1]{{\frame{\includegraphics[height=60pt,width=82pt]{figs/frame_uti/#1}}}}
\begin{figure}[h!]
\centering
\resizebox{.99\linewidth}{!}{
\begin{tabular}{cccc} 
Image & Ground Truth & FlowNetS & FlowCaps-S\\
\UTITH{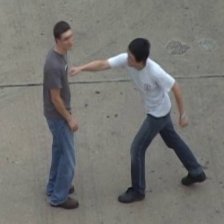} & \UTITH{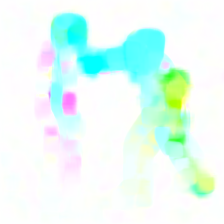} & \UTITH{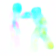} & \UTITH{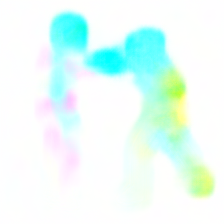}\\
\UTITH{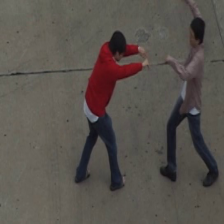} & \UTITH{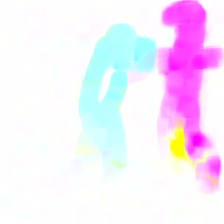} & \UTITH{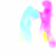} & \UTITH{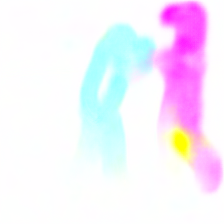}\\
\UTITH{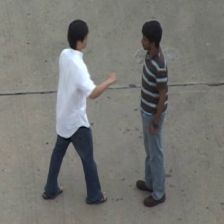} & \UTITH{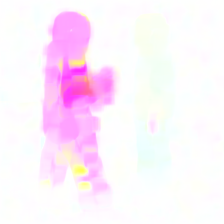} & \UTITH{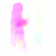} & \UTITH{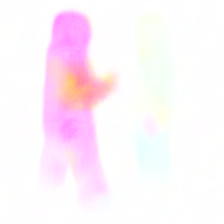}\\
\end{tabular}
}
\caption{The optical flow estimation results on the UTI P-Frames dataset.}
\label{fig:frame_uti}
\end{figure}

\begin{figure*}[t]
\centering
\includegraphics[scale=0.4,trim={4cm 0.5cm 4cm 0},clip] {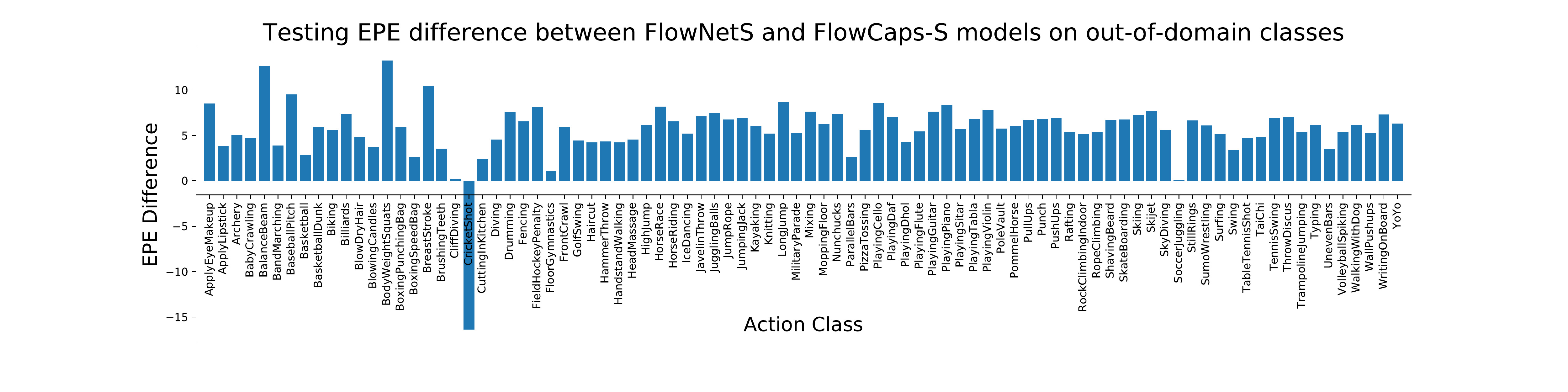}
\caption{Testing EPE value differences between the FlowNetS and FlowCaps-S model performances on the out-of-domain action classes of the UCF I-frames dataset.}
\label{fig:ood}
\end{figure*}

\begin{figure}[t]
\begin{tabular}{cc}
\includegraphics[width=0.5\columnwidth]{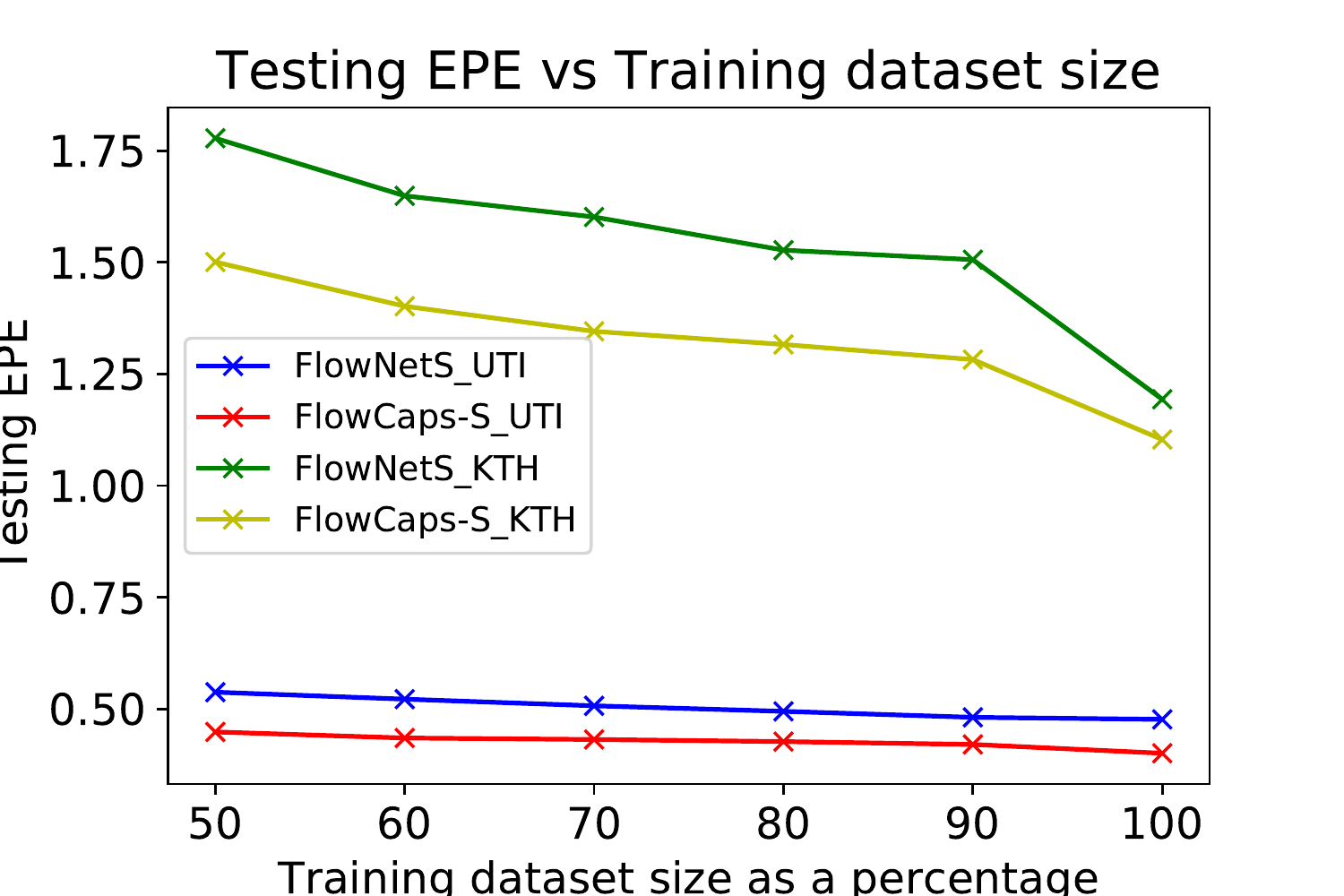} & \includegraphics[width=0.5\columnwidth]{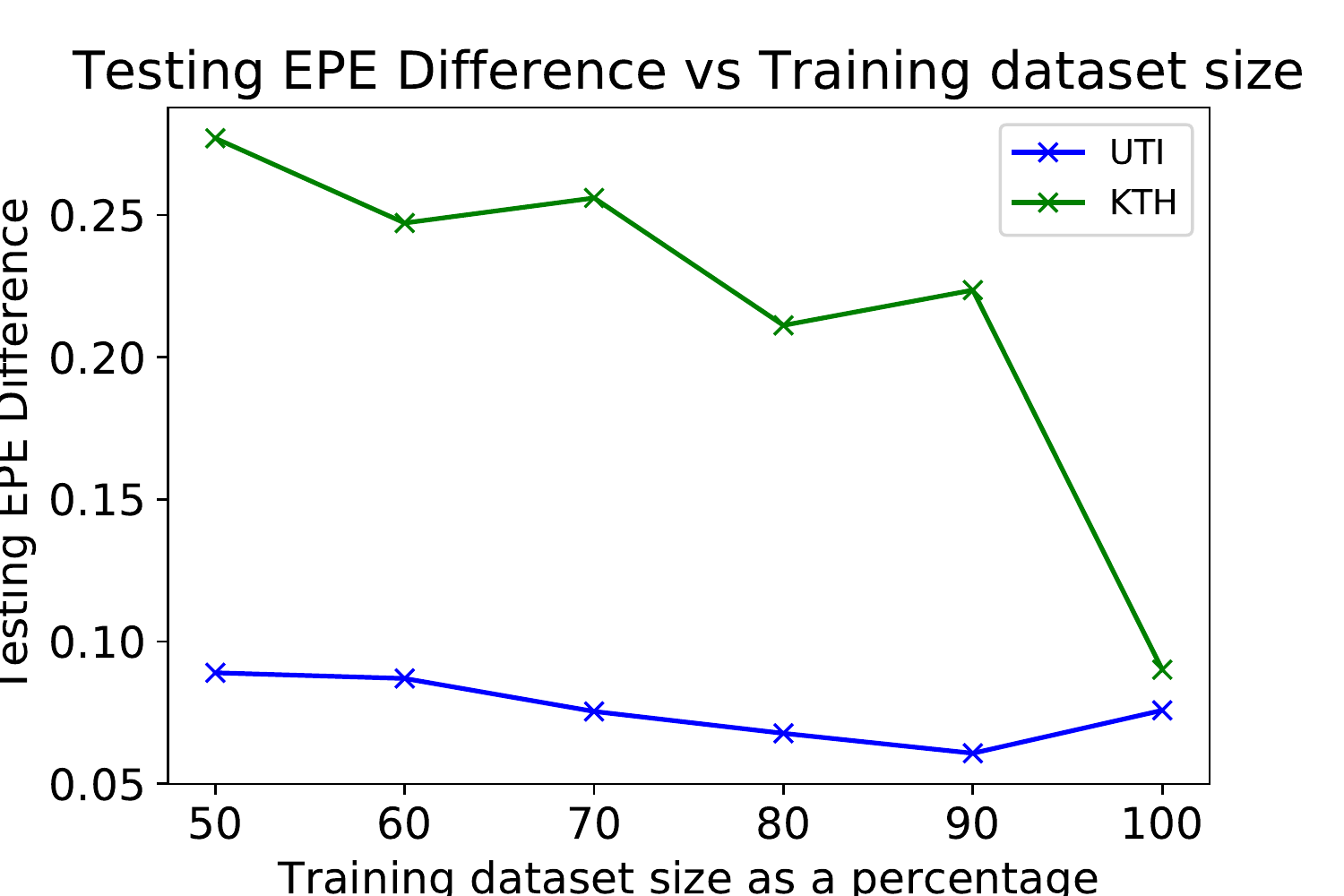}\\
(a)&(b)
\end{tabular}
\caption{The comparison of testing EPE values achieved by the FlowNetS and FlowCaps-S models: (a) Testing EPE vs Training dataset size; (b) Testing EPE difference vs Training dataset size.}
\label{fig:lowdata_epe}
\vspace{-4mm}
\end{figure}

Table \ref{tbl:epe} compare our results to that of FlowNetS obtained for frame-wise and segment-wise  optical flow estimation.
On the KTH I-frames dataset, the proposed FlowCaps-S model outperformed the FlowNetS model by a significant margin of 7.55\%, by achieving an average testing EPE value of 1.1033.
The relative improvement obtained for segment-wise is even significant where our model outperforms FlowNetS by 17.33\% obtaining 0.938 EPE.
The optical flows estimated by the two models in comparison to the ground truth flows are illustrated by Fig. \ref{fig:frame_kth}. 

A similar result was observed on the Sub UCF I-Frames dataset as the proposed FlowCaps-S model achieved an average testing EPE value of $2.2037$, after outperforming the FlowNetS model performance by a notable margin of $4.80\%$.
A visual inspection of generated optical flow shown in Fig. \ref{fig:frame_ucf} indicates the superiority of our method. 
Preserving the trend, the FlowCaps-S model outperformed the FlowNetS model by a comprehensive margin of $14.01\%$ while achieving an impressive average testing EPE value of $0.3806$. Fig. \ref{fig:frame_uti} illustrates the optical flows estimated by the two models on the UTI P-Frames dataset. 
Hence, it was evident that across all the datasets, the proposed FlowCaps-S model outperformed the FlowNetS model for the optical flow estimation task. Furthermore, it is interesting to note that, the percentage improvement is inversely proportional to both the complexity of the dataset and the number of training samples, suggesting that the proposed CapsNet-S model better generalizes with less number of training data, and on less complex datasets, in comparison to the FlowNetS model.
Most interestingly, segment-wise model consistently outperform frame-wise model indicating the advantage of our 3D convolution-based capsule encoder and better exploitation of contextual motion information.
Both FlowNetS and FlowCaps-S benefit from segment-wise model, however the improvements obtain by our model is better than the FlowNetS.

As shown in Table~\ref{tbl:epe}, for action classification task from the estimated optical flows, the results follow the same pattern as the optical flow estimation task, where the flows estimated with the proposed FlowCaps-S model outperformed those of the FlowNetS model by significant margins. 
Quantitatively, the proposed FlowCaps-S model contributed to achieving $65\%,91.20\%$ and $86.02\%$ on the KTH-I Frames, Sub UCF I-Frames and UTI-P Frames datasets respectively, while outperforming its counterpart model by respective margins of $3.70\%,5.50\%$ and $1.90\%$. 
Similar trends can be seen for segment-wise model.
Hence, it can be concluded that the optical flows estimated by the proposed FlowCaps-S model can be better-adopted to other tasks such as action classification, than those by the FlowNetS model.

Furthermore, it is interesting to observe that ground truth action recognition performance is sometimes lower than our model.
However, we do not expect this behavior on more challenging datasets.
Our model is a learning-based optical flow estimation method. 
Essentially, the model learns about motion information through entire dataset and hence able to capture dataset specific biases as well.
Therefore, we hypothesise that good optical flow learning models might be able to exploit motion information and biases in a dataset and may be able to output flow images that contain more motion information suitable for action recognition.
Furthermore, segment-wise approach with a 3-dimensional architecture outperforms the frame-wise approach for both optical flow estimation and action classification tasks across almost all datasets. 
We conclude that perhaps segment-wise approach is better for optical flow estimation and action recognition.

\subsection{Other strengths of FlowCaps}
FlowCaps is able to generalize to out-of-domain using fewer training samples. Here, we consider the I-frames extracted from the entire UCF-101 dataset. We test with both models on all the classes of UCF-101 except for classes with no videos containing more than 5 I-frames, and for the five classes considered for training in the Sub UCF I-Frames dataset, which yields $88$ out-of-domain action classes. 

Fig. \ref{fig:ood} illustrates the differences in the testing EPE values obtained from the FlowNetS and FlowCaps-S models. It is evident from observation that except for the \textit{Cricket Shot} action class, the proposed FlowCaps-S model achieves a lower testing EPE than the FlowNetS model in rest of the 87 out-of-domain classes, suggesting that the proposed FlowCaps-S model generalizes to out-of-domain optical flow estimation better than the FlowNetS model.

Our model learns well with a low amount of training data. 
We hypothesize that when the fraction of the training dataset used decreases, the difference between the average testing EPEs of  FlowNetS and FlowCaps-S models should increase, indicating better generalization of the proposed FlowCaps-S model with less training data.
To this end, we train the models with fractions of the training data ranging from $50-100\%$ with $10\%$ intervals for the KTH-I Frames and UTI P-Frames datasets, and plot the raw EPE values and the corresponding differences in Fig. \ref{fig:lowdata_epe} (a) and Fig. \ref{fig:lowdata_epe} (b) respectively. It is evident from observation that except for the UTI P-Frames dataset instance with the full training set, the rest of the instances favor our hypothesis, across both datasets. Hence, we concluded that the FlowCaps-S model generalizes better than the FlowNetS model for optical flow estimation, with less training data. 

\section{Conclusion}
In this paper we have investigated a deep model for optical flow estimation by extending FlowNetS~\cite{dosovitskiy2015flownet} and Capsule Networks~\cite{sabour2017dynamic} coined FlowCaps-S. We investigated frame-based model and a video segment-based model that utilizes 3D convolutions. 
We consistently outperform several state-of-the art models for both optical flow estimation on classical benchmarks and on some important action recognition datasets.
Interestingly, our model have only a fraction of parameters compared to other baselines.
We demonstrate that our model is able to learn with few examples and generalize to out-of-domain examples better than other counterparts.

\section*{Acknowledgment}
\noindent
This research/project is supported by the National Research Foundation, Singapore under its AI Singapore Programme (AISG Award No: AISG-RP-2019-010).

{\small
\bibliographystyle{ieee_fullname}
\bibliography{main}
}

\end{document}